\documentclass{article}

\usepackage{arxiv}

\usepackage[utf8x]{inputenc} % allow utf-8 input
\usepackage[T1]{fontenc}    % use 8-bit T1 fonts
\usepackage{hyperref}       % hyperlinks
\usepackage{url}            % simple URL typesetting
\usepackage{booktabs}       % professional-quality tables
\usepackage{amsfonts}       % blackboard math symbols
\usepackage{nicefrac}       % compact symbols for 1/2
\usepackage{microtype}      % microtypography
\usepackage{lipsum}
\usepackage{xcolor}
\usepackage{graphicx}
\usepackage{setspace} 
\usepackage{amsmath}
\usepackage{array}
\usepackage{mdwmath}
\usepackage{mdwtab}
\usepackage{eqparbox}
\usepackage{url}
\usepackage{multirow}
\usepackage{textcomp}

\title{DNNSurv: Deep Neural Networks for Survival Analysis Using Pseudo Values}

\author{
  Lili Zhao\\
  Department of Biostatistics\\
  University of Michigan\\
  Ann Arbor, MI 48109 \\
  \texttt{zhaolili@umich.edu} \\
   \And
  Dai Feng \\
  Biometrics Research Department\\  Merck Research Laboratories\\ Kenilworth, NJ 07033 USA \\
  \texttt{fengdai73@gmail.com} \\
}

\begin{document}
\maketitle

\doublespacing

\begin{abstract}
There has been increasing interest in modelling survival data using deep learning methods in medical research. Current approaches have focused on designing special cost functions to handle censored survival data. We propose a very different method with two simple steps. In the first step, we transform each subject's survival time into a series of \textit{jackknife} pseudo conditional survival probabilities and then use these pseudo probabilities as a quantitative response variable in the deep neural network model. By using the pseudo values, we reduce a complex survival analysis to a standard regression problem, which greatly simplifies the neural network construction. Our two-step approach is simple, yet very flexible in making risk predictions for survival data, which is very appealing from the practice point of view. The source code is freely available at \url{http://github.com/lilizhaoUM/DNNSurv}
\end{abstract}

% keywords can be removed

\section{Introduction}

Recently, using deep neural networks to predict when an event of interest will happen has gained
considerable attention. These studies are often characterized by incomplete observations, in
particular right-censored data; e.g. patients may be lost to follow-up without experiencing the
event. Thus, handling the censored data becomes the crucial aspect of these analyses.

\cite{Martinsson2016} developed a neural network assuming that survival time has a Weibull
distribution. This parametric assumption is too restrictive in the analysis of large datasets. The
semi-parametric Cox proportional hazards regression is the most widely used method for analyzing
censored survival data. It studies the effects of the covariate variables on survival by estimating
hazard functions, and it assumes that the hazard ratio for any two subjects is constant over time;
that is, the proportional hazards (PH) assumption. Several authors have adapted the Cox model to
the deep neural networks \cite{Katzman:2018,Ching:2018,Zhu:2016}. They trained the deep neural
network models by minimizing the negative partial likelihood defined under the PH assumption.
However, this assumption is often questionable when the number of covariates is large, as every
covariate needs to satisfy the PH assumption.

Another school of deep neural network modelling for survival data utilized the discrete-time
survival model, including recent papers by \cite{Gensheimer:2018, Fotso:2018,
Lee:2018,Luck:2017,Giunchiglia:2018}. In this modelling framework, the follow-up time is divided
into a set of fixed intervals. For each time interval, the conditional hazard/probability is
estimated: the probability of failure in the interval, given that the individual has survived up to
the beginning of the interval. Compared to the Cox model, the discrete-time survival model is more
flexible as it does not rely on the PH assumption.  However, the censored data in each interval are handled either by  an
oversimplified method that assumes individuals with a censoring time in the second half of an
interval survived that interval \cite{Gensheimer:2018}, or by an over complicated method that uses a
ranking loss function \cite{Lee:2018} or a combination of a ranking loss and an isotonic regression
\cite{Luck:2017}.

All existing deep neural network models referenced above require the development of a special cost
function to handle the censored data, sometimes with a sophisticated network structure, such as a
convolutional or recurrent neural network \cite{Zhu:2016,Giunchiglia:2018}. In this article, we
develop a very different approach. To circumvent the complexity introduced by censored data, we
substitute the observed survival times by \textit{jackknife} pseudo observations, and then use
these pseudo observations as a quantitative response variable in a regression analysis powered by
the deep neural networks.

Summary of nice features in our proposed method: 
\begin{itemize}
    \item[1.]  Compared to the PH-based neural network survival models, it  outputs a survival probability, which is often of direct interest to patients and physicians rather than a hazard ratio, or prognostic index, in the PH-based methods.
    \item[2.] Compared to existing discrete survival models, it uses a theoretically justified method to deal with the censored data, which is based on the established theory in the survival analysis using pseudo-observations.
    \item[3.] Unlike all existing neural network survival models,  it uses a simple conventional loss function, which can be easily adapted to various datasets.
\end{itemize}

The remainder of this article is organized as follows. In section \ref{method}, we present our
two-step method and address the covariate-dependent censoring problem. In section \ref{sims}, we
conduct simulation studies to evaluate our model and demonstrate its  superior performance  over the existing methods. In section \ref{realdata}, we apply neural network models to three real datasets and show that our model outperforms the PH-based models when the PH assumption is violated. We conclude this paper
with a brief discussion in section \ref{discuss}.

\section{METHOD}
\label{method}

\textit{Notation.} Let $X_i$ be the survival time for subject $i$, $C_i$ be the censoring time,
$T_i=\min(X_i,C_i)$ be the observed survival time, and $\delta_i=I(X_i \leq C_i)$ be the censoring
indicator. We assume that survival times and censoring times are independent. Let
$\mathbf{Z_i}=(z_{i1},\cdots, z_{ip})$ denote the $p$-dimensional covariates. We first assume that
the censoring distribution does not depend on the covariates and then address the
covariate-dependent problem in section \ref{dc}.

\subsection{Review the current pseudo-observations approach.}
\label{review}
 The pseudo-observations approach
\cite{Andersen:2003,Klein:2005,Klein:2007,Andersen:2010} provides an efficient and straightforward
way to study the association between the covariates and survival outcome in the presence of
censoring.

If the outcome of interest is the survival probability at specific time $t$, without incomplete
data, we could directly model $I(T_i> t )$ on $\mathbf{Z}_i$ ($i=1,\cdots,n$) using a generalized
linear model (GLM) with a \textit{logit} link, for a binary outcome variable. In the presence of
censoring, $I(T_i> t)$ is not observed for all subjects. In this case the Kaplan-Meier (KM)
estimator can be used to estimate the survival probability at any given time point. The KM
estimator is approximately unbiased under independent censoring \cite{Andersen:1993}, which is a
requirement for the validity of the pseudo-observations approach. Based on the \textit{jackknife}
idea, a pseudo survival probability is computed for each (censored and uncensored) subject. For the
$i^{th}$ subject, the pseudo survival probability is computed by

\begin{align}
\widehat{S}_i(t) &= n \widehat{S}(t) - (n-1) \widehat{S}^{-i}(t),
\label{pseudo}
\end{align}
\noindent where  $\widehat{S}(t)$ is the KM estimator of $S(t)$ using all $n$ subjects and
$\widehat{S}^{-i}(t)$ is the KM estimator using sample size of $n-1$ by eliminating the $i^{th}$
subject. Then  $\widehat{S}_i(t)$ ($i=1,\cdots,n$) are used as a numeric response variable in the
standard regression analysis, which is similar to model fit to $I(T_i > \tau ),$ $(i=1,\cdots,n),$
if these values were observed.

\cite{Klein:2005,Klein:2007,Klein:2008} proposed computing a vector of pseudo survival
probabilities at a finite number of time points equally spread on the event time scale for each
subject, and then modelling these pseudo survival probabilities as a function of the covariates by
the generalized estimating equation (GEE), with the \textit{complementary log-log} link, which is
equivalent to fitting a Cox model to the survival data. The regression coefficient estimates from
this pseudo-based GEE is approximately consistent when the censoring distribution is independent of
the covariates and of the survival times \cite{Andersen:2003}.

\subsection{Compute pseudo conditional probabilities.}

In this article, we adapt the  pseudo-observations approach in a discrete-time survival framework.
We first divide the follow-up time into $J$  intervals: $(0,t_{1}]$,$(t_1,t_2]$,$\cdots,$
$(t_{J-1},t_J]$. For each interval, we compute the pseudo conditional survival probability: the
probability of surviving the interval, given that the subject has survived the previous interval.
For a given interval $(t_{j},t_{j+1}]$, all subjects who are still at risk is denoted by $R_{j},$
and the pseudo conditional probability of surviving $t_{j+1}$ is computed as
\begin{equation}
\hat{S}_{ij}(t_{j+1}|R_j) = R_j \hat{S}(t_{j+1}|R_j)-(R_j-1) \hat{S}^{-i}(t_{j+1}|R_j),
\label{pseudos}
\end{equation}

\noindent where $\widehat{S}(t_{j+1}|R_j)$ is the KM estimator constructed using the remaining
survival times for all patients still at-risk at time $t_j,$ and $\widehat{S}^{-i}(t_{j+1}|R_j)$ is
the KM estimator for all patients at-risk but the $i^{th}$ subject. For the first interval
$(0,t_{1}],$  all subjects are at risk (i.e., $R_0=n$). If there is no censored data, the pseudo
probability in each interval is either 0 or 1. With censoring, the pseudo probability is a real
value; that is, it can be above 1 or below 0; see properties of the pseudo observations discussed
in \cite{Andersen:2010}.

By using the discrete-time survival framework, we transform each subject's observed survival time
(censored or uncensored) into a series of pseudo conditional survival probabilities. Unlike the
pseudo-based GEE, in which the pseudo probabilities are computed from the marginal survival
function (i.e., defined from time zero), the pseudo probabilities in our approach are computed from
the conditional survival function. Therefore, our pseudo probabilities are conditionally independent and we do not need to consider the within-subject correlation as in the pseudo-based GEE model.

Choosing the time intervals needs to be done carefully. It has been shown that as few as five time
intervals equally spread on the event time scale worked quite well in most cases \cite{Klein:2007}.
One could also divide the time based on time points of clinical relevance; for example, patients'
prognosis at 5 and 10 years might be of particular interest to the investigators. In either case,
the interval can not be too small or too large. If the interval is too large, data information is
lost. Conversely, if the interval is too small, no or a few events would occur in the interval,
leading to an inefficient analysis. As fewer subjects remain in the study at later follow-up time,
the upper bound of the last interval, $t_J$, should be reasonably smaller than the maximum
follow-up time, so that we have sufficient information in the last interval.

We created a function in R  called \textit{getPseudoConditional} to compute the pseudo conditional
probabilities described above. The R code is available in the GitHub repository
\url{http://github.com/lilizhaoUM/DNNSurv/blob/master/getPseudoConditional.R}.

Table \ref{exp} is an example output for the  three hypothetical subjects. The Pseudo Probability in the last column is the pseudo conditional survival probability for a subject at a particular time $t$. Each subject has multiple pseudo probabilities, one for each time point, and the number of pseudo probabilities may vary as different subjects might have different follow-up times. For example, subject 1 is not at risk at 18 months because this subject had the event or was lost to followup before this time, so there is no pseudo probability at 18 months or later. The outputs from the \textit{getPseudoConditional} function are then used as inputs in a deep neural network model.

\begin{figure}
\centering
 \includegraphics[height=5.5cm, width=8cm]{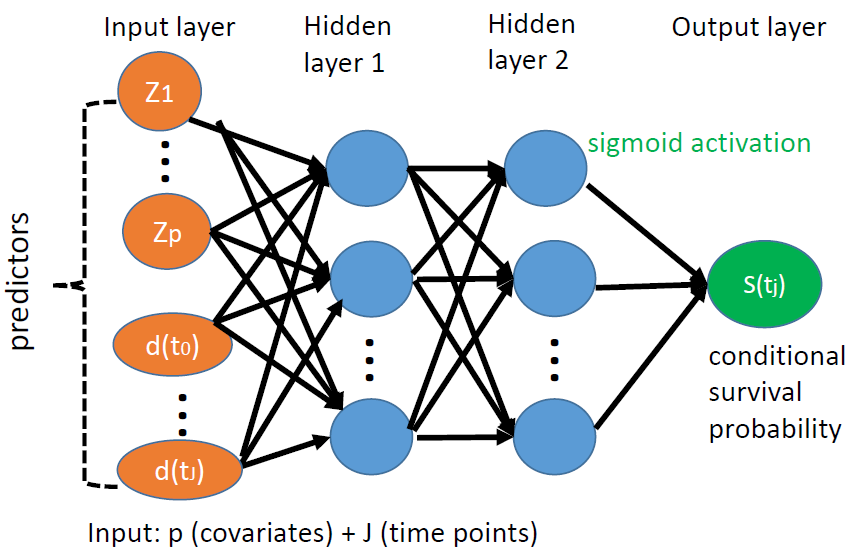}
\caption{DNNSurv Architecture with two fully connected hidden layers.}
\label{arch}
\end{figure}

{The input predictor variables can be $z_1$  and $t$, resulting in 1 (covariate) $+$ 1 (time) input nodes. Alternatively, we can covert the continuous time $t$ into a categorical variable by creating an indicator variable for each category. For example, there are 4 indicator variables, d(0), d(6), d(12) and d(18) for time 0, 6, 12 and 18, respectively. In this case, there are 1 (covariate) $+$ $J$ (time points) input nodes. We found that the alternative approach (i.e., indicators for $t$) was generally better than the continuous $t$. Therefore, we have used this network structure in both simulation studies and real data analysis.

%\newcolumntype{g}{>{\columncolor{Gray}}c}
\begin{table}[!t]
\caption{An example output from \textit{getPseudoConditional}. }
\label{exp}
\centering
\begin{tabular}{|c|c|c|c|c|c|c|c|}
\hline
\multirow{2}{*}{ID}  &  \multirow{2}{*}{$z_1$} &  \multirow{2}{*}{\textcolor{gray}{t}} &\multicolumn{4}{|c|}{Indicators for $t$: } & Pseudo \\\cline{4-7}
& & & d(0) & d(6) & d(12) & d(18)  & Probability \\
\hline
1	& 3.2& \textcolor{gray}{0}& 1 & 0 & 0 & 0  &	1.039   \\
1	& 3.2&	\textcolor{gray}{6}& 0 & 1 & 0 & 0  & 	1.017  \\
1	& 3.2&	\textcolor{gray}{12}&0 & 0 & 1 & 0  	&-0.014   \\
2	& 5.8&	\textcolor{gray}{0}& 1 & 0 & 0 & 0 & 	1.039  \\
2	& 5.8&	\textcolor{gray}{6}& 0 & 1 & 0 & 0 & 	1.017  \\
2	& 5.8&	\textcolor{gray}{12}&0 & 0 & 1 & 0 &  	1.025 \\
2	& 5.8&	\textcolor{gray}{18}&0 & 0 & 0 & 1 &  	0.726  \\
3	& 1.5&	\textcolor{gray}{0}& 1 & 0 & 0 & 0 & 	1.039    \\
3	& 1.5&  \textcolor{gray}{6}& 0 & 1 & 0 & 0 &  	1.017\\
3	& 1.5&	\textcolor{gray}{12}&0 & 0 & 1 & 0 &  	1.025 \\
3	& 1.5&	\textcolor{gray}{18}&0 & 0 & 0 & 1 &  	1.080\\
\hline
\end{tabular}
\end{table}

\subsection{Architecture of our proposed network}
We have named our pseudo-value based deep neural network model  DNNSurv.  Figure \ref{arch} shows an architecture of DNNSurv  with two fully connected hidden layers.} The output of the network is a single node, which predicts the conditional survival probabilities at a particular time point. To constrain the probability between 0 and 1, the sigmoid activation function is used for the final layer.

By using the Pseudo Probability in Table \ref{exp} as the response variable, the complex survival data analysis is reduced to a regression analysis with a single quantitative response variable.  Thus, we can directly train DNNSurv by using the conventional cost function, which minimizes the mean of squared differences between pseudo conditional survival probabilities and predicted conditional survival probabilities.  We implemented DNNSurv in the Keras library \cite{chollet2015keras} in R with Tensorflow \cite{tensorflow2015-whitepaper} backend (code is available at \url{http://github.com/lilizhaoUM/DNNSurv}).

\subsection{Compute survival probabilities from DNNSurv}

Given a set of $p+J$ input values, DNNSurv is able to predict the conditional survival probability
in each interval, that is, $P(T > t_{j}| T > t_{j-1})$ for $j=1,\cdots, J$. The marginal survival
probability, $P(T > t_{j}),$ is calculated by multiplying the conditional survival probabilities up
to the $j^{th}$ interval:

$$P(T > t_j)= \prod_{k=1}^j P(T > t_{k}| T > t_{k-1}).$$

Thus, the discrete-time survival framework allows us to predict both the marginal  and the
conditional survival probability, or the complementary risks. Both the marginal and conditional
estimates have important clinical implications. For example, a patient who is diagnosed with lung
cancer might be interested in the probability of surviving one year (i.e., the marginal survival
probability). If the patient has survived the first year, then he/she might be interested in the
probability of surviving another year (i.e., the conditional survival probability).

\subsection{Handling covariate dependent censoring}
\label{dc}

KM estimates used in creating the pseudo values are subject to covariate-dependent censoring bias.
In this case, we propose to use the inverse of probability of censoring weighted (IPCW) estimator
for the survival function, denoted by $\hat{S}^W(t)$, which has been successful at reducing the
bias \cite{Binder:2014,Xiang:2012}). We replace $\hat{S}(t)$ by $\hat{S}^W(t)$ in (\ref{pseudos})
to compute the IPCW pseudo conditional survival probabilities by
\begin{equation}
\hat{S}^W_{ij}(t_{j+1}|R_{j}) = R_{j} \hat{S}^W(t_{j+1}|R_{j}) - (R_{j}-1) \hat{S}^{{W}^{-i}}(t_{j+1}|R_{j}),
\label{pseudosw}
\end{equation}

\noindent where $\hat{S}^W(t) = \exp \{-\hat{\Lambda}^W(t)\},$ and $\hat{\Lambda}^W(t)$ is the IPCW
Nelson-Aalen estimator for the the cumulative hazard function and is estimated by

$$\hat{\Lambda}^W(t)=\sum^n_{i=1} \int^t_0 \frac{d N_i(u)  \hat{W_i}(u)}{\sum^n_{j=1} Y_j(u)\hat{W_j}(u)},$$

\noindent where $N_i(u)=I(T_i \leq u, \delta_i=1)$ is the observable counting process for subject
$i,$ $Y_j (u) = 1(T_j \geq u)$ is the at risk process for subject $j$, and $\hat{W_i}(u)$ is the
inverse of probability of censoring for subject $i$ at time $u$. By assigning different weights for
subjects based on their covariate values, $\hat{S}^W(t)$ is approximately unbiased if the censoring
distribution is correctly specified \cite{Binder:2014}. Calculation of the IPCW pseudo conditional
survival probabilities is also implemented in the \textit{getPseudoConditional} function.

To model the censoring time distribution,  we consider the Cox model, but other models, such as
accelerated failure time model (AFT) \cite{wei1992} or deep neural networks, could work as well. In
calculating the pseudo probabilities, we fit the censoring model once and then use the same
censoring model for all subjects; that is, $\hat{W_i}(u)$ ($i=1,\cdots,n$) remain the same in
computing the pseudo probabilities in (\ref{pseudosw}). Then we replace the pseudo probabilities with the IPCW  pseudo probabilities in DNNSurv to predict the survival probability. We refer this model as DDNSurv{\_}ipcw in this article.

\subsection{Comparison to existing neural network model}
\label{other}
In both simulation studies and real data analyses, we compared our DNNSurv to three existing neural network survival models, which include two PH-based neural network models, Cox-nnet \cite{Ching:2018} and DeepSurv \cite{Katzman:2018}, and one recently published discrete-time neural network survival model, nnet-survival \cite{Gensheimer:2018}. These three models represent two schools of neural network survival models.  Moreover, these methods have undergone peer review and have well documented source codes.

\begin{table*}[!t]
\caption{Comparison to existing neural network models. }
\label{compare}
\footnotesize
\begin{tabular}{|l|l|l|l|l|}
\hline
Parameters	&	DNNSurv	&	nnet-survival	&	DeepSurv	&	Cox-nnet\\
\hline
Assumption &	no	&	no 	&	PH	&	PH\\ \hline
\# input nodes	&	p (covaraites) + J (indicators)	&	p (covaraites)	&	p (covaraites)	&	p (covaraites)\\ \hline
Response 	&	numeric pseudo probabilities  	&	failure/at-risk indicators 	&	time/censoring indicators	&	time/censoring indicators\\ \hline
Output	&	survival probability	&	survival probability	&	log hazard ratio	&	log hazard ratio\\ \hline
Cost function	&	sum of squared errors	&	discrete model likelihood 	&	Cox partial likelihood	&	Cox partial likelihood\\ \hline
Censoring data	&	pseudo-value method	&	location in the interval 	&	Cox model	&	Cox model\\ \hline
\end{tabular}
\end{table*}

Table \ref{compare} shows detailed comparisons among the four neural network models. DNNSurv and nnet-survival do not rely on the PH assumption. Compared to nnet-survival, DNNSurv uses a theoretically justified pseudo-value approach to deal with censored data, whereas nnet-survival arbitrarily determines a subject to be a survivor or not depending on whether  the censoring occurs in the first or second half of the interval. We have also proposed a strategy to handle data with covariate-dependent censoring, which seems to be the only limitation in DNNSurv. On the contrary, nnet-survvial has no mechanism to handle the dependent censoring.

\section{SIMULATIONS}
\label{sims}

We conducted simulation studies to compare our DNNSurv to existing neural network models, as described in Table \ref{compare}.

\subsection{Evaluate the model performance}

To evaluate the model performance, we considered two metrics commonly used in survival analysis. The first metric was the time-dependent concordance index (c-index) \cite{Harrell:1996}, which
measures  how well a model predicts the ordering of sample event times. The other metric was the
Brier score \cite{Gerds:2006}, which evaluates the accuracy of a predicted survival probability at
a given time point by measuring the average squared distances between the observed survival status
and the predicted survival probability, with a smaller value indicating a better performance. 

Cox-nnet and DeepSurv both output a log hazard ratio ($\hat{\theta}_i$) rather than the survival probability. To compute the survival probability at a time point, we first
computed the Nelson-Aalen estimate of the baseline survival function, $\hat{S}_0(t)$, and then
computed the survival probability for subject $i$ by
$\hat{S}_i(t)=\hat{S}_0(t)^{\exp(\hat{\theta}_i)}$.

\subsection{Determine hyperparameters in neural network model}

\label{hyper}
We determined the hyperparameters in the above neural network models using a random Cartesian grid search \cite{bergstra2012}. The hyperparameters include number of hidden layers (1 or 2 layers), number of nodes (4, 8, 16, 32, 64, or 128 nodes), dropout regularization \cite{Srivastava:2014}  (drop rate of 0.2 or 0.4) or  ridge regularization \cite{Ching:2018} (penalty of 0.0001, 0.001 or 0.01), activation function, and optimization algorithm with a learning rate of 0.001, 0.005 or 0.01. For the activation function and optimization algorithm, we only considered those that were discussed in the published paper \cite{Ching:2018,Katzman:2018,Gensheimer:2018}.  Cox-nnet is optimized to work on high dimensional gene expression data and includes the optimization for some hyperparameters (such as dropout rate and learning rate) within the package, which makes it difficult to change parameters for the network structure; for example, no example illustrates  how to change the number of layers. Therefore, we used the defaults for some parameters  (e.g., Nesterov optimizer and one hidden layer). See details in the Supplementary Table S1 for the hyperparameters used in each network model for each study.

To determine the best set of hyperparameters,  we used 5-fold CV with c-index as the performance metric on the training set. Once we had determined the best set of hyperparameters, we trained the model again using all training data and predicted the survival probability on the test data.

\subsection{Simulations from an AFT model}
\label{simind}

Survival data were generated from an AFT model. Since the deep neural network model is able to
approximate complex nonlinear functions, we considered the flexible random function generator in
\cite{friedman2001} for the mean function in the AFT model \cite{henderson2017}. For all of the
studies presented in this subsection, the number of variables  is $20$ (i.e., $\mathbf{Z}=(z_1,
\ldots, z_{20})$ ), and their joint distribution follows a standard multivariate normal
distribution. The nonlinear mean function in the AFT model takes the form

\begin{center}
\begin{align} \mu(\mathbf{Z})=\sum_{l=1}^{10} a_l g_l(\mathbf{Z_{(l)}}), \label{sim}
\end{align}
\end{center}

\noindent where coefficients $a_l$'s were generated from a uniform distribution $a_l \sim U[-1,1]$.
Each $g_l$ is a Gaussian function of a randomly selected variables, $\mathbf{Z_{(l)}},$ of the $20$
variables; see details in \cite{friedman2001} on the selection of these variables and the Gaussian
function construction. The expected number of variables for each $g_l$ function was 4. By using a
linear combination of 10 $g_l$ functions,  the mean in the AFT model is a function of all, or
nearly all, of the $20$ variables with different strength of association with the survival outcome,
and it also involves higher-order interactions between some of the variables. Finally, we generated
the residuals in the AFT model from a gamma distribution with a shape parameter of 2 and a rate
parameter of 1, resulting in a signal-to-noise ratio of 3.

We simulated 100 datasets, and each dataset has a different mean function, $\mu(\mathbf{Z}),$
generated from (\ref{sim}). The censoring times were independently generated from an exponential
distribution. A different rate parameter was chosen to obtain a censoring rate of 0.2, 0.4, or 0.6,
which corresponds to light, moderate and heavy censoring, respectively. We generated 5000
observations, 75\% of which were randomly chosen as training data and the remaining 25\% were test
data. As each simulation produced a different dataset, we divided the time period into six intervals from the $10^{th}$ to the $60^{th}$ percentile of the empirical survival distribution, which avoided the complication of zero events in any interval caused by using the same set of time points for all datasets.

\begin{figure}
  \centerline{\hbox{ \hspace{0.0in}
    \includegraphics[width=8cm,height=6cm]{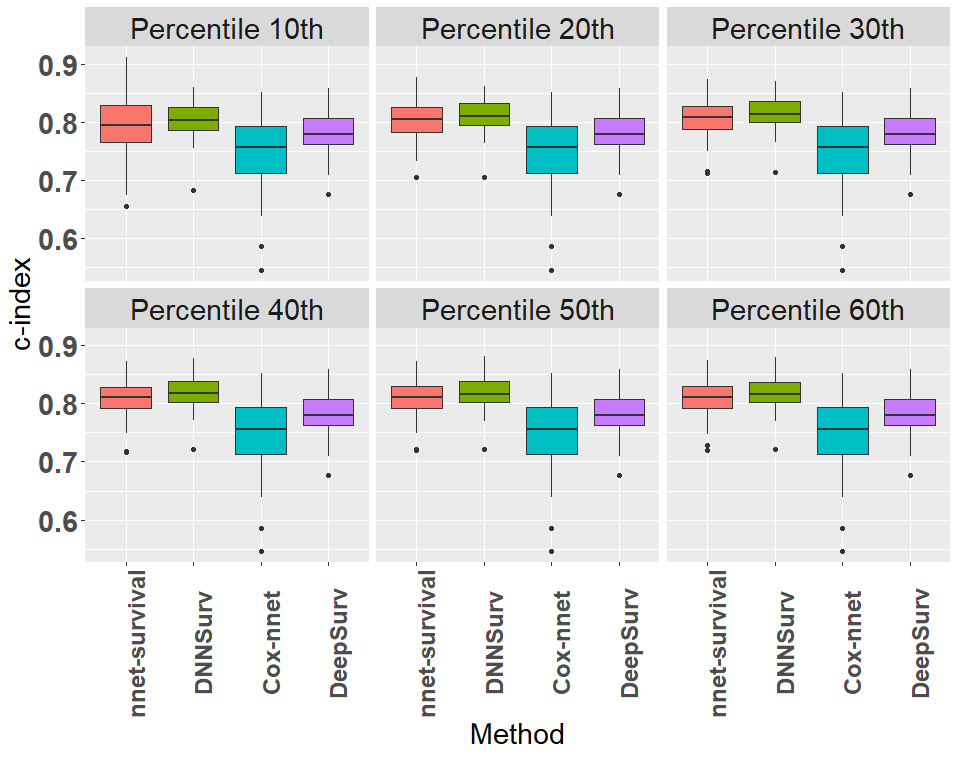}
    \includegraphics[width=8cm,height=6cm]{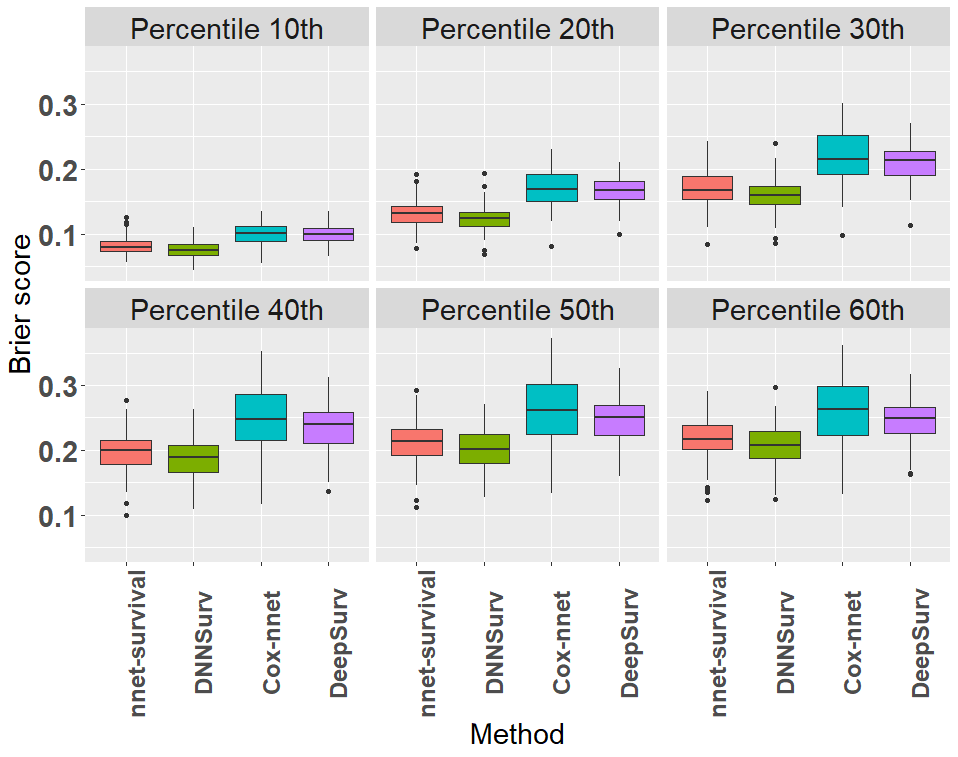}
    }
  }
\caption{Boxplots of c-index (left) and Brier score (right) from  DNNSurv, DeepSurv, Cox-nnet, and nnet-survival over 100 simulated 
datasets generated from the AFT model with Friedman's random function generator with 40\%  censoring. The c-index and Brier scores were evaluated at six time points, which were determined from the six percentiles of the empirical survival distribution. }
\label{fig:cindex}
\end{figure}

Figure \ref{fig:cindex} shows the c-index and Brier scores from the four neural network models. Both DNNSurv and nnet-survival outperformed the DeepSurv and Cox-nnet, which indicates the advantage of the  discrete-time models over the PH-based models when data are generated from non-PH models. DNNSurv had similar, or slightly better performance than nnet-survival.  Similar conclusions were drawn for datasets with 20\% and 60\% censoring, with results  shown in Supplementary Figure 1S.

We  performed a sensitivity analysis to investigate the influence of the number of intervals on the prediction accuracy. We reduced the 6 intervals into 2 intervals at the $20^{th}$ and $40^{th}$ percentiles of the empirical survival distribution. We found that the c-index and Brier scores remained almost the same (boxplots  shown in Supplementary Figure 2S), which indicates that the results are fairly robust to the choice of number of time intervals.

\subsection{Simulations with covariate dependent censoring}

To investigate the  IPCW estimator, we set up a simple simulation study with covariate dependent censoring. In this study,
survival data were generated from a Cox model with one covariate, $z,$ $\lambda(t|z)=0.1 \exp(\beta z),$ where
$\beta=1$ and $z$ was simulated from a standard normal distribution. The censoring model followed the same Cox model for the survival times, resulting in approximately $50\%$ censoring rate in each simulated dataset. In each simulation, we randomly selected $2000$ subjects to train the model and predict survival probabilities for a separate set of $2000$ subjects at the $10^{th},$ $20^{th},$ $30^{th},$ $40^{th},$ and $50^{th}$ percentiles of the overall survival distribution.

\begin{figure}
  \centerline{\hbox{ \hspace{0.0in}
    \includegraphics[width=8cm,height=7cm]{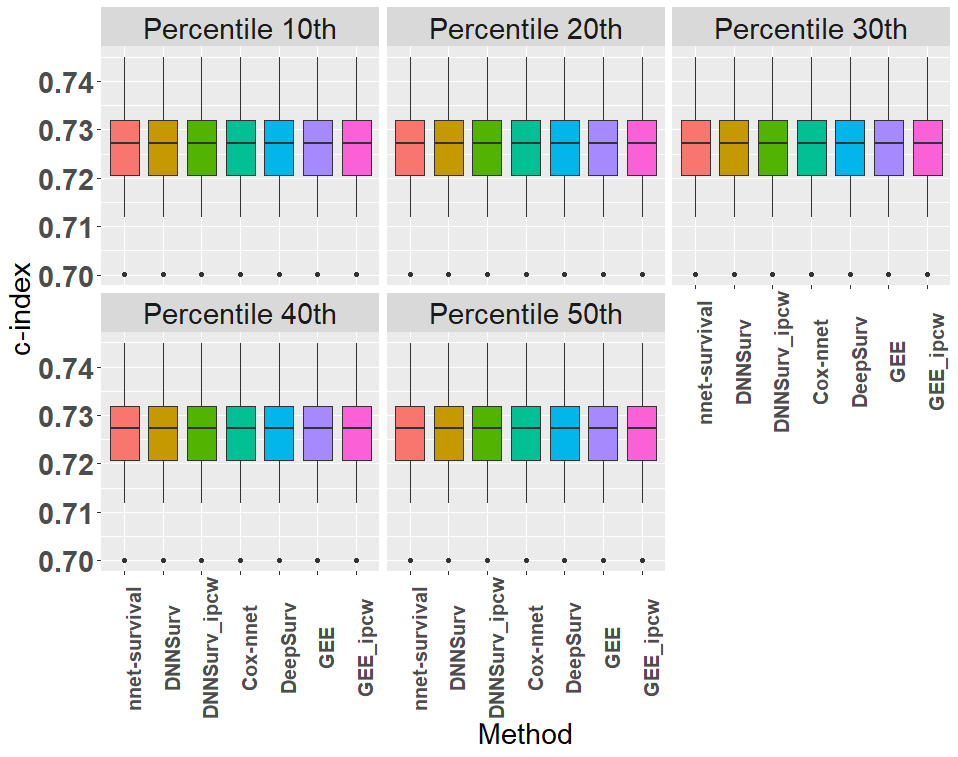}
    \includegraphics[width=8cm,height=7cm]{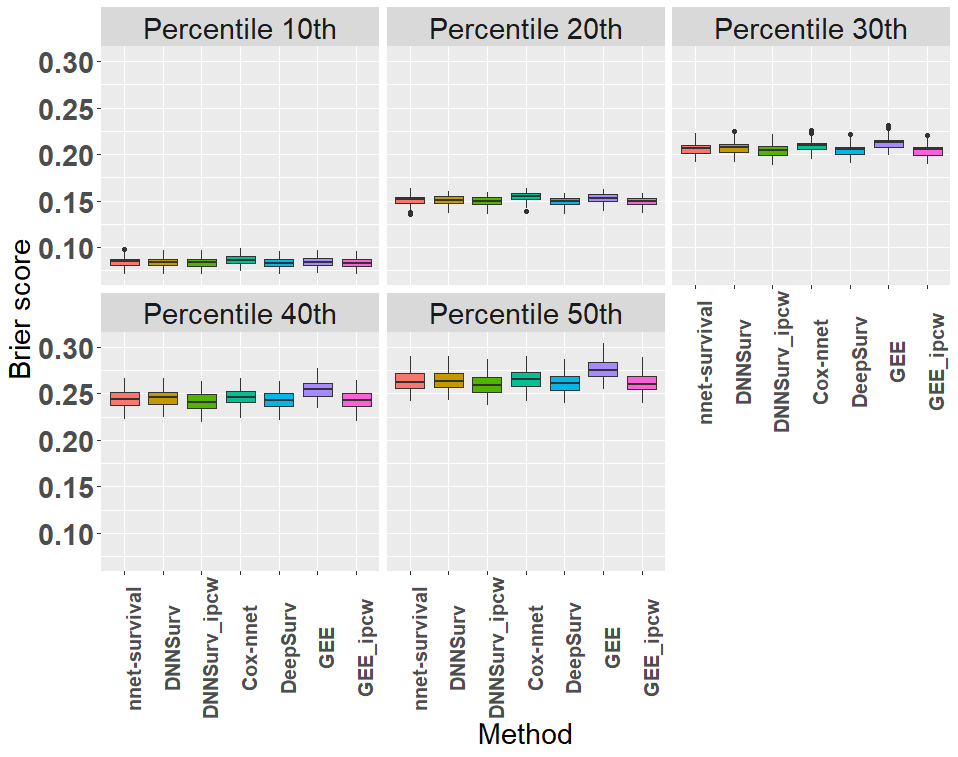}
    }
  }
\caption{Boxplots of c-index (left) and Brier score (right) for the 100 simulated datasets generated from the simple Cox model in the case of covariate-dependent censoring.}
\label{exp1wc}
\end{figure}

We first applied the GEE method as described in Section \ref{review}, which is analogous to the standard Cox model (see details in \cite{Andersen:2003,Klein:2007}).  We then applied the GEE with the IPCW pseudo values (denoted by GEE{\_}ipcw). Based on 100 simulations, we found that GEE produced biased estimates for parameter $\beta$ (mean is 0.783 and  mean squared error (MSE) is 0.05), while  GEE{\_}ipcw reduced the bias (mean is 1.002 and  MSE is 0.0026). Thus,  the IPCW method is efficient in correcting the bias in the GEE model. Finally, we applied five neural network models: DDNSurv, DDNSurv{\_}ipcw, Cox-nnet, nnet-survival, and DeepSurv to the simulated data.  Figure \ref{exp1wc} shows boxplots of the c-index over 100 simulations. Surprisingly, all the studied methods have the same c-index, possibly because the rank-based c-index measure is not sensitive to the bias when the covariate effect is monotonic on survival.

The accuracy measure,  Brier scores in Figure \ref{exp1wc}, shows various findings: 1) GEE had the worst performance due to the bias, whereas GEE{\_}ipcw had the best performance as it corrected the bias, and the model matched the true data generating model; 2) compared to GEE,  neural network models were less sensitive to the violation of the independent censoring assumption; 3) DNNSurv{\_}ipcw improved the prediction accuracy over the DNNSurv, and it had similar performance as DeepSurv and GEE{\_}ipcw; and 4) Cox-nnet had the worse performance as the data were generated under the PH assumption. The reason might be that Cox-nnet is optimized to work on high dimensional gene expression data, and it might not work well when there is only one predictor.

\section{REAL APPLICATION}
\label{realdata}

\textbf{CHS data.} The Cardiovascular Health Study (CHS) was initiated in 1987 to determine the risk factors for
development and progression of cardiovascular disease (CVD) in older adults, with an emphasis on subclinical measures. Detailed description of the study can be found in \cite{DrPhilos:1993}. The event of interest was time to CVD. The study has collected a large number of variables at baseline, including demographics (e.g., age, gender and race), family history of CVD, lab results and medication information, with the goal of identifying important risk factors for the CVD event. We selected 29 predictor variables to make predictions of the CVD event; see a complete description of the variables in Table S2. After excluding subjects with missing data in any of the selected predictor variables, we had $5,380$ subjects,  65.2\% of whom had CVD during the study period. We predicted the survival probability in every year up to 15 years. In the CHS data, an interval of one year is not small, as the event rate  is large and reasonable numbers of the event were observed in each interval.

\textbf{MESA data.} The second CVD study we considered was the Multi-Ethnic Study of Atherosclerosis (MESA) study, which
enrolled subjects who were free from clinical cardiovascular disease from six communities in the
United States in 2000$-$2002. Participants were followed for identification and characterization of
cardiovascular disease events. Detailed description of the study can be found in \cite{Bild:2002}.
Similar to the CHS study,  the event of interest was time to CVD.  We  selected 30 variables to
make predictions of the CVD event; see a complete description of the predictor variables in Table
S4. After excluding subjects with missing data in any of the selected predictor variables, we had
$6,547$ subjects, 5.4\% of whom had CVD during the study period. The event rate is  small,
an interval of one year was to short. Thus, we
predicted the survival probability only at two clinically meaningful landmark
time points: 3 and 5 years.

\textbf{SRTR data.} Data were obtained from the Scientific Registry of Transplant Recipients (SRTR). The SRTR dataset includes patients who underwent kidney transplantation between 2011 and 2013.  Participants were followed from transplantation to graft failure or death, whichever occurred first.
Based on published literature \cite{He:2019}, we considered covariates age, race, gender, BMI,  donor height, cold ischemic time, indicator of anti-viral therapy and immunosuppressant therapy for the prediction of kidney failure.  After excluding subjects with missing data in any of the selected predictor variables, we had
$7,288$ subjects, 11\% of whom  experienced kidney failure during the study period.  We predicted the survival probability at four time points: 1 day, 1 month, 6 months, 8 months and 1 year.

In addition to the four network models (DNNSurv, nnet-survvial, DeepSurv and Cox-nnet), we also  applied a standard Cox proportional hazards model (CPH) \cite{Cox:1992}. Compared to DeepSurv and Cox-nnet,  CPH models the log hazard ratio as a linear function of covariates (implemented using the coxph function in R). In each study, we randomly selected 75\% subjects as training data and the other 25\% as test data, and repeated 10 times.

Table \ref{res} shows the averaged c-index for each of the five models. DNNSurv had similar performance as the PH-based models (CPH, DeepSurv and Cox-nnet) in both CHS and MESA studies, which indicates that the PH assumption is reasonable in both datasets. It is unclear why nnet-survival had a poor performance in the MESA study, possibly due to its \textit{ad-hoc} method of handing the censored data. In the SRTR study, DNNSurv and nnet-survival had significantly better performance than the PH-based models. A test of the PH assumption (by cox.zph function in R) reveals that this assumption was strongly violated for several variables in this dataset, including cold ischemic time,  anti-viral therapy and immunosuppressant therapy. All models had the same Brier scores, so those results are not presented here.

\begin{table}[!t]
\renewcommand{\arraystretch}{1.3}
\caption{c-index from five models in three real studies}
\label{res}
\centering
\begin{tabular}{|c|c|c|c|c|c|}
\hline
\textbf{Study} & \textbf{CPH}	&	\textbf{DNNSurv}	&	\textbf{nnet-survival}	&	\textbf{DeepSurv}	&	\textbf{Cox-nnet}\\
\hline
\textbf{CHS}	& 0.70	&0.69	&	0.69	&	0.70  & 0.70\\ \hline
\textbf{MESA} & 0.73	&0.73	&	0.69	&	0.73  & 0.73\\ \hline
\textbf{SRTR} & 0.72	&0.78	&	0.77	&	0.72  & 0.74\\ 
\hline
\end{tabular}
\end{table}

We also tried DNNSurv{\_}ipcw in the CHS study with weights calculated based on four variables that were significantly associated with the censoring time using the Cox model. Model performances were the same as the DNNSurv.

\section{CONCLUSION}
\label{discuss}

In this article, we develop a two-step approach for making risk predictions in survival analysis
using a deep neural network model. We first compute the \textit{jackknife} pseudo survival probabilities in the discrete-time survival framework, and then substitute the survival times by these pseudo probabilities to make risk predictions in a deep neural network model.  The IPCW pseudo
probabilities are also proposed in case of the covariate-dependent censoring. By using the pseudo values,
the analysis for censored survival data is reduced to a regression problem with a quantitative
response variable, which greatly facilitates the use of deep learning methods. Standard deep neural
networks can be directly applied, which avoids the difficulty of designing a special cost function
for the censored data, as in the current methods.

We demonstrated the superior performance of DNNSurv over existing methods in both simulation studies and real data analysis when the PH assumption is violated.  DNNSurv is also a competitive alternative to the PH-based neural network models when the data satisfies the PH assumption. Compared to the PH-based neural network models,  DNNSurv directly outputs survival probabilities, which are often of direct interest to patients and physicians, rather than the hazard ratio in the PH-based models.

\section{Acknowledgements}

This Manuscript was prepared using the CHS and MESA data obtained from the National Heart, Lung,
and Blood Institute (NHLBI) Biologic Specimen and Data Repository Information Coordinating Center
and does not necessarily reflect the opinions or views of the NHLBI.CHS and MESA data were obtained
from the National Heart, Lung, and Blood Institute (NHLBI) Biologic Specimen and Data Repository
Information Coordinating Center (\url{https://biolincc.nhlbi.nih.gov/}). Thanks to Kirsten Herold for proofreading the article.
%%%%%%%%%%%%%%%%%%%%%%%%%%%%%%%%%%%
%%                               %%
%% Additional Files              %%
%%                               %%
%%%%%%%%%%%%%%%%%%%%%%%%%%%%%%%%%%%
\section*{Additional Files}
  \subsection*{Additional Table S1 --- Hyperparameters in Cox-nnet and DNNSurv}
  \subsection*{Additional Table S2 --- Description of variables in CHS study}
  \subsection*{Additional Table S3 --- Brier scores for DNNSurv in the CHS data }
  \subsection*{Additional Table S4 --- Description of variables in MESA study}

\bibliographystyle{unsrt}  
\bibliography{btc}

\end{document}